%% file: paycheck.tex
\documentclass[11pt]{article} 
\usepackage[left=1in,right=1in,top=1in,bottom=1in]{geometry}

\input{math_commands.tex}

\usepackage{hyperref}
\usepackage{url}
\usepackage{float}

\title{Reinforcement Learning Paycheck Optimization for Multivariate Financial Goals}
\author{Melda Alaluf 
\and
Giulia Crippa\textsuperscript{\ref{princetonORFE}}
\and
Sinong Geng\textsuperscript{\ref{princetonCS}} 
\and
Zijian Jing
\and
Nikhil Krishnan\textsuperscript{\ref{princetonORFE}}
\and
Sanjeev Kulkarni\textsuperscript{\ref{princetonORFE},\ref{princetonECE}}
\and
Wyatt Navarro
\and
Ronnie Sircar\textsuperscript{\ref{princetonORFE}}
\and
Jonathan Tang}

\usepackage{bbm}
\usepackage{bm}
\usepackage{mathtools}
\usepackage{amsmath}
\usepackage{xcolor}

\usepackage{dirtytalk}

\newcommand{\curly}[1]{\left\{#1\right\}}

\usepackage{amsthm}
\usepackage{dsfont}

\newcommand{\EE}{\mathbb{E}}

\usepackage{mathtools}
\DeclarePairedDelimiter\abs{\lvert}{\rvert}%
\makeatletter
\let\oldabs\abs
\def\abs{\@ifstar{\oldabs}{\oldabs*}}

\usepackage{makecell}
\usepackage{natbib}
\usepackage{graphicx}

\usepackage{booktabs} 
\usepackage{multirow}
\usepackage{float}

\usepackage{pbox}

\usepackage{graphicx}

\usepackage{capt-of}
\usepackage{caption}
\usepackage{subcaption}
 
\usepackage{microtype}
\usepackage{wrapfig}

\usepackage{algorithm}
\usepackage{algorithmic}

\usepackage{comment}

\newcommand{\eq}[2]{\begin{equation} \label{eq:#1} #2 \end{equation}}
\newcommand{\eqs}[1]{\begin{equation*} #1 \end{equation*}}

\usepackage[inline]{enumitem}
\newlist{inlineenum}{enumerate*}{1}
\setlist*[inlineenum,1]{%
      label=(\roman*),%
}

\newsavebox{\tempboxa}
\newsavebox{\tempboxb}
\newsavebox{\tempboxc}

\newlist{todolist}{itemize}{2}
\setlist[todolist]{label=$\square$}
\usepackage{pifont}

\usepackage{color}
\usepackage{scalerel}
\usepackage[symbol]{footmisc}

\usepackage{makecell}

\usepackage[capitalize,noabbrev]{cleveref}

 \newtheoremstyle{theorem}{6pt}{6pt}{\rm}{}{\sffamily}{ }{ }{}
 \theoremstyle{theorem}

 \newtheoremstyle{lemma}{6pt}{6pt}{\rm}{}{\sffamily}{ }{ }{}
 \theoremstyle{lemma}

 \newtheoremstyle{assumption}{6pt}{6pt}{\rm}{}{\sffamily}{ }{ }{}
 \theoremstyle{assumption}
 
\newtheoremstyle{case}{6pt}{6pt}{\rm}{}{\sffamily}{. }{ }{}
 \theoremstyle{case}

 \newtheoremstyle{statement}{6pt}{6pt}{\rm}{}{\sffamily}{ }{ }{}
\theoremstyle{statement}

 \newtheoremstyle{corollary}{6pt}{6pt}{\rm}{}{\sffamily}{ }{ }{}
 \theoremstyle{corollary}
 
  \newtheoremstyle{definition}{6pt}{6pt}{\rm}{}{\sffamily}{ }{ }{}
 \theoremstyle{definition}

\newtheoremstyle{example}{6pt}{6pt}{\rm}{}{\sffamily}{ }{ }{}
\theoremstyle{example}

\newtheoremstyle{remark}{6pt}{6pt}{\rm}{}{\sffamily}{ }{ }{}
\theoremstyle{remark}

\newtheoremstyle{approximation}{6pt}{6pt}{\rm}{}{\sffamily}{ }{ }{}
\theoremstyle{approximation}

\newtheoremstyle{scheme}{6pt}{6pt}{\rm}{}{\sffamily}{ }{ }{}
\theoremstyle{scheme}

\newtheoremstyle{Algorithm}{6pt}{6pt}{\rm}{}{\sffamily}{ }{ }{}
\theoremstyle{Algorithm}

\newtheoremstyle{Assumption}{6pt}{6pt}{\rm}{}{\sffamily}{ }{ }{}
\theoremstyle{Assumption}

\newtheoremstyle{proposition}{6pt}{6pt}{\rm}{}{\sffamily}{ }{ }{}
\theoremstyle{proposition}

\newtheoremstyle{hypo}{6pt}{6pt}{\rm}{}{\sffamily}{ }{ }{}
 \theoremstyle{hypo}

  \newtheoremstyle{Step}{6pt}{6pt}{\rm}{}{}{ }{ }{}
 \theoremstyle{Step}

\begin{document}

\maketitle
\begingroup\renewcommand\thefootnote{\fnsymbol{footnote}}
\footnotetext[1]{ORFE Department, Princeton University\label{princetonORFE}, Princeton, NJ 08544, USA}
\footnotetext[4]{Computer Science Department, Princeton University\label{princetonCS}, Princeton, NJ 08544, USA}
\footnotetext[3]{Electrical \& Computer Engineering Department, Princeton University\label{princetonECE}, Princeton, NJ 08544, USA}
\endgroup

\begin{abstract}

We study paycheck optimization, which examines how to allocate income in order to achieve several competing financial goals.
For paycheck optimization, a quantitative methodology is missing, due to a lack of a suitable problem formulation. 
To deal with this issue, we formulate the problem as a utility maximization problem.
The proposed formulation is able to \begin{inlineenum}
    \item unify different financial goals;
    \item incorporate user preferences regarding the goals;
    \item handle stochastic interest rates.
\end{inlineenum}
The proposed formulation also facilitates an end-to-end reinforcement learning solution, which is implemented on a variety of problem settings.

\end{abstract}
\section{Introduction }
We propose a reinforcement learning solution to paycheck optimization.
Specifically, one aims to allocate monthly income in order to achieve goals like paying out loans, purchasing a mortgage, saving for retirement, etc. 
Indeed, such a problem is common in everyday life and similar services are provided by various companies.
In this work, we hope to provide a rigorous framework for such problems with a reinforcement learning solution.

Finding a suitable problem formulation for paycheck optimization is especially challenging. 
First, the goals of paycheck optimization are often multivariate and quite heterogeneous. 
Therefore, it is cumbersome to unify such goals and optimize them simultaneously. 
Second, incorporating the preferences of users becomes especially complicated for paycheck optimization. 
For instance, some goals like paying out credit card debt can be more urgent to the user than saving money or purchasing a mortgage, while the amounts vary greatly.
It is unclear how to incorporate such information into decision-making methods. 
Third, the interest rates of financial goals (e.g. inflation rate, savings rates, etc.) evolve stochastically over time. 
This makes learning an optimal paycheck allocation strategy even more challenging. 
Finally, without a proper formulation, the powerful decision-making tools in machine learning and control are not applicable to this problem. 

To the best of our knowledge, a quantitative solution for paycheck optimization is missing.
Existing results on paycheck optimization are mainly analytical without an implementable methodology~\citep{swart2004personal,archuleta2011future,hershey2013effective}.  
One can also consider paycheck optimization as a non-traditional robo advising problems with different targets.
However, existing literature is mainly for portfolio optimization~\citep{giudici2022network,d2019promises,d2021robo} or other single financial goals \citep{d2021new}: such methods are not applicable to paycheck optimization with the multiple heterogeneous goals studied in this work. 
Some existing paycheck optimization solutions rely on a simple waterfall method. 
Specifically, the user needs to prioritize different goals in an absolute order to finish the goals one by one. 
In other words, all incomes will be allocated to a specific goal and only when one is met will the next one be considered.
As a result, the method is incapable of targeting multiple goals simultaneously and
thus is generally sub-optimal. 
An example where the waterfall method performs poorly is given in Appendix \ref{app_waterfall}.

Separately, there exists a huge amount of literature on reinforcement learning for decision-making in various scenarios. 
Such methods provide flexible solutions for many different decision-making problems, but are not directly applicable to paycheck optimization. 

In this work, we propose a utility maximization framework for paycheck optimization and a data-driven policy gradient method.  
First of all, we formulate the paycheck optimization as a utility maximization problem to unify various financial goals and incorporate user preferences. 
In detail, we leverage piecewise-linear utility functions. 
Whenever a goal is active (i.e. it is still beneficial to allocate income to this goal) the corresponding utility function is negative, while it becomes zero otherwise. 
This design has two advantages: on one hand, it encourages a policy to finish the goals; 
on the other hand, it is possible to express the user-specific preference for each goal via
the slope of the utility function - the steeper the slope, the more beneficial it is to allocate income to the corresponding goal.

The decision-making target is to maximize the sum of the utility functions of each goal across time.
With the proposed utility maximization framework, we conduct policy gradient to solve for an optimal paycheck allocation strategy.
Specifically, with the collected data, we implement policy learning using gradients estimated from the data. 
As a result, we learn a paycheck optimization policy in a data-driven and model-free manner, without specifying any stochastic model. 
This provides a flexible solution to paycheck optimization~\footnote{
The implementation of the method is provided via \url{https://github.com/gengsinong/paycheck_optimization}
}.

\section{Problem Formulation}
Paycheck optimization studies the problem of income allocation over different financial goals. 
Examples of financial goals include paying out credit card debt, paying out student loans, saving for a home down payment,  saving emergency funds, or saving for retirement using for example 401Ks or IRAs. 
At time $t$, we use $S_t$ to denote a user's income, and $\pi_t$ to denote the fraction of $S_t$ assigned to different financial goals. 
By optimizing the income allocation $\pi_t$, we aim to complete all the financial goals. 

In paycheck optimization, the financial goals are heterogeneous. For
instance, savings goals like retirement and emergency funds are different from debt goals like student
loan and credit card debt. Indeed, the former depend on interest rates that increase the value of the
wealth assigned to them, i.e. they contribute to finishing them. On the other hand, the latter’s interest rates
increase the value of the debt itself, delaying their completion. Goals also differ based on their
maturity. Short-term goals, like credit card debt, have much higher interest rates and thus, should be
treated more urgently with respect to longer-term goals. In practice, it is unclear how to unify such
heterogeneous traits and optimize these goals simultaneously.

While the main objective of paycheck optimization is finishing all financial goals, it is also important to consider the users’ preferences. Instead of completing the goals as fast as possible, different users may have different priorities for each goal. For example, some users could prefer saving for purchasing a house more than saving for retirement. On the other hand, some might prioritize retirement, given that such investments may have higher interest rates that would allow them to finish all other goals more quickly. Additionally, an agent may choose to pay down high interest debt first, or they may prioritize the confidence boost that comes from zeroing out debt and choose to pay out small debts first. As a result, how to quantify the preference of users and incorporate them into paycheck optimization is also an open question.

\section{Paycheck Optimization as Utility Maximization}
In this section, we formulate paycheck optimization as a utility maximization problem. 
For each financial goal, we define \begin{inlineenum}
\item a state variable, 
\item its dynamics and,
\item a utility function.
\end{inlineenum}

\subsection{State variable}
To devise a utility maximization objective, we first define the state variable for this problem. 
Let $I$ be the set of financial goals with cumulative totals that we aim to achieve. 
Then, $X_t \in \mathbb{R}^{|I|}$ is the state variable, i.e., the fraction of each goal that still needs to be paid out at time point $t \in \{0,...,T\}$.
For $i \in I$, the component $X_t^i$ gives a normalized measure of the proportion of goal $i$ that is left to complete, where $X_t^i=0$ denotes that the goal has been completed by time $t$, and $X_0^i=1~\forall i \in I$ since we start off without having contributed income towards any of the goals.

\subsection{Dynamics}
\label{sec:update}
Critically, the specific dynamics of $X_t$ are different for each financial goal.
We use $S_t$ and $\pi_t^i $ to denote the monthly income (increasing at the rate of inflation) and the fraction of income assigned to goal $i$ at time $t$ respectively.  
For goals involving debt repayment like credit card debt and student loans, the dynamics of $X_t^i$ is given by
\eq{dynamic_debt}{
X_{t+1}^i  = (1+r^i_t)X_{t}^i -  \frac{S_t \pi_t^i}{G^i}\textnormal{, with }i \in \{ \text{Credit Card Debt, Student Loans} \}
}
where $r^i_t$ is the interest rate for goal $i$ and $G^i$ is the total amount for goal $i$.
In words, $X_t^i$ is updated according to the interest rate, before subtracting the proportion of the goal that the user will pay at time $t$.
In practice, the interest rates and goal amount are different for each financial goals (see Table~\ref{tab:inputs}).


For financial goals involving savings like home down payment and emergency funds, we define
\begin{align}
    X^i_{t+1}=1-(1+r_t^i)(1-X^i_t)-\frac{S_t \pi_t^i}{G^i }\textnormal{, with } i \in \{ \text{Home Down Payment, Emergency Funds} \}
\end{align}
which differs from the previous case, since the interest rate contributes to completing the different goals. In general, the emergency fund will not have an interest rate, since it is intended to be easy and accessible cash on-hand.  However, we assume that mortgage savings are invested in a risk-free financial instrument with some constant rate of return, while in Section~\ref{sec:extension} we will consider stochastic rates for this investment.

Finally, for retirements savings, the structure is considerably different due to the presence of tax-advantaged savings accounts which are commonly used for retirement, i.e., 401K and IRA.  If goal $i$ is the retirement savings, denoted as $RS$, then
\begin{align*}
    X^{\text{RS}}_{t+1}=1-(1+r_t^{\text{RS}} )(1-X^{\text{RS}}_t)-\frac{\displaystyle m_t+S_t^{RS}(\pi^{401K}_t+\pi^{IRA}_t+\pi^{RS}_t)}{G^{\text{RS}} }
\end{align*}
where $r^{\text{RS}}$ is the rate of returns on retirement savings (which can again be extended to evolve stochastically).  Note that $(1-X^{\text{RS}}_t)$ is the fraction of retirement savings which has been already paid out, and so $1-(1+r^{\text{RS}}_t )(1-X^{\text{RS}}_t)$ gives the fraction of debt still outstanding after applying the rate of return. Eventually, we subtract the amount paid to the retirement goal, as well as the amount paid to 401K and IRA (since they also contribute to retirement). Also, we deduct a factor $m$ which represents employer matching for 401K, which depends on the level defined in the employer's specific plan.

\subsection{Utility}
For each goal we wish to define a corresponding utility function $u_i: \mathbb{R} \rightarrow \mathbb{R}$.
We define the utility function as a function of the fraction finished in each financial goal, denoted as $\bar{X}_t^i = 1-{X}_t^i$.
To formulate the priority of task $i$, specified by users, we use two positive constants $p^i$ and $q^i$, where a larger $p^i$ (or $q^i$) corresponds to a greater urgency to complete goal $i$.
Specifically,  with $\bar{X}_t^i = \bar{x}$, $p^i=p$, $h^i=h$, and $q^i = q$, we define two types of utility functions:
\begin{align*}
    w_1(x; p)&=-p\cdot \max(0,1-\bar{x}) \\
    w_2(x; p,q, h)&=-q\cdot \max(0,1-\bar{x})-(p-q)\cdot \max(0,1-\bar{x}-h),
\end{align*}
where $w_1$ represents the utility for a single-phase goal, while $w_2$ a two-phase one. 
This allows us to specify multiple priorities $p$ and $q$, depending on the proportion of goal already been met. 
Note that $w_2$ includes a parameter $h \in [0,1]$ to specify the crossover point between when a user pays off the first segment of the goal and moves onto the second one. Below we give a plot of such utility functions. 

\begin{figure}[H]
    \centering
    \includegraphics[scale=0.43]{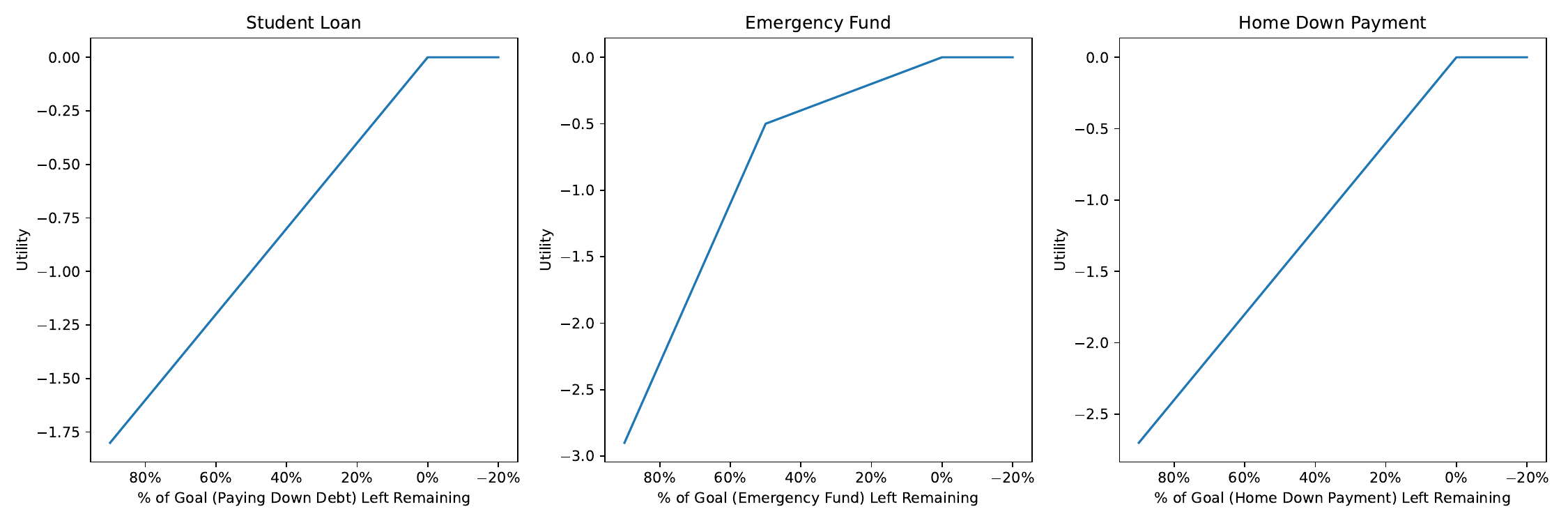}
    \caption{Different utility functions}
    \label{fig:utility2}
\end{figure}

Let us now note the following about the defined utility functions.
First, both are continuous with
finite derivative, which allows for gradient-based methods to solve the corresponding maximization problem \eqref{eq:target} below.  
Second, they return negative values when $x>0$, but stay at zero when $x\leq 0$. Moreover, as stated above, the value of $p,q$ for different financial goals model the user's preference with respect to each goal. Specifically, $p,q$ are non-negative weights which can be chosen to incentivize completion of goals: the larger the value of $p$, the more incentive to complete the first (or only) segment of the goal; the larger the value of $q$, the greater the incentive to complete the second segment. 

For the single phase goals, e.g. $i \in$ \{Credit Card Debt, Student Loans, Home Down Payment, Retirement Savings\}, we have
\begin{align*}
    u_i(X_t^i)=w_1(X^i_t; p^i),
\end{align*}
where $p^i$ is the weight for goal $i$.  For the emergency fund we have
\begin{align*}
    u_{\text{EF}}(X_t^{EF})= &w_2\bigg(X^{\text{EF}}_t; p^{\text{EF}}, q^{\text{EF}},h^{EF} \bigg),
\end{align*}
where $(1-h^{EF})G^{EF}$ and $h^{EF}G^{EF}$ are the amounts to be completed with urgency $p^{EF}$ and $q^{EF}$ respectively.

In addition to the goals with state variables, we also want to assign utilities to the 401K and IRA goals, since making regular contributions to these accounts could be a user desired goal in addition to their contribution to the retirement savings goal.  Since there are no cumulative totals to meet, the utility for the 401K and IRA will be assessed based on the contribution at each time $t$.

401K is a two-phase goal, where the first phase consists of contributions up to a minimum level, while the second one comprises contributions up to the maximum allowed level. Hence, for 401K the utility will be given by
\begin{align*}
    u_{401K}(X_t^{401k})= &w_2\bigg(1-\frac{\pi^{\text{401K}}_t}{M^+ }; p^{401K},q^{401K},\frac{M^+-M^- }{ M^+ }\bigg),
\end{align*}
where $M^+,M^-$ are the maximum and minimum income percentage contribution levels to 401K respectively. For IRA instead,
\begin{align*}
    u_{\text{IRA} }(X^{IRA}_t)=w_1\bigg(1-\frac{S_t \pi^{\text{IRA}}_t}{I^+ } ; p^{\text{IRA} } \bigg),
\end{align*}
where $I^+$ is the maximum permitted income contribution to the IRA. \\ 

Eventually, with utilities defined for every and each goal, we can define the paycheck optimization target as maximizing the expected total utility:  
\eq{target}{
\max_{\pi_t} V(\pi_t) \quad \textnormal{ with } \quad V(\pi_t) = \sum_{t=0}^{T}\sum_{i \in I} u_i(X_t^{i,\pi}) , 
}
where the evolution is over the dynamics of $X_t^{i,\pi}$, under policy $\pi_t$.

Note that, while the proposed framework provides the ability to assign a different preference (weights) to every goal, it is always feasible to fix some of them to simplify the problem. 
This is vital from a practical perspective, since it might be difficult for users to order the importance of all financial goals. 
For instance, we might assume that the user have the same preference for all debt goals and thus
fix $p^i$ for student loans and credit card debt as one value.
In practice, this can simplify the communication with users when trying to come up with $p^i$'s and $q^i$'s.

\section{Deep Deterministic Policy Gradient}
\label{sec:method}
We aim to solve \eqref{eq:target} by deep deterministic policy gradient~\citep{silver2014deterministic}. 
Specifically, we parametrize the policy as a deep neural network of $X_t$:
\eq{policy}{
\pi(t) = f(X_t; \theta),
}
where $f$ denotes the neural network with parameter $\theta$.
Then, for each iteration, we update the parameter $\theta$ by the estimated gradients of $V(\pi_t)$ with respect to $\theta$.\\ Our procedure is as follows:

\begin{algorithm}[H]
    \caption{Paycheck optimizer}
  \begin{algorithmic}[1]
    \STATE {\bfseries Input:} Amounts for each goal, interest rates, matching percents, number of iteration $N$.
    \FOR{$n \in [N]$}
    \STATE Initialize $X_0$ as one for each goal.
    \STATE For each month $t$:
    \begin{enumerate}[label=(\alph*)]
        \item $\pi_t=f(X_t; \theta) $
        \item Update $X_t$ from $X_{t-1}$ according to the rules of Section~\ref{sec:update}
    \end{enumerate}
    \STATE Estimate $\frac{d V(\pi_t)}{d \theta}$
    \STATE Maximize $V(\pi_t)$ with respect to $\theta$
    \ENDFOR
    \STATE {\bfseries Output:} Return $\theta$
  \end{algorithmic}
\end{algorithm}

Where Step 1 is initialization for each training epoch, Step 4(b) follows \eqref{eq:policy}, and Steps 5,6 are where we optimize our neural network.  Optimization can be done with any standard algorithm, for example gradient descent or ADAM~\citep{kingma14}.

\section{Simulation with Constant Rates}
\label{sec:exp-deter}
In this section, we implement the proposed method for paycheck optimization. 
We aim to show that our method is readily available for paycheck optimization for users with different preferences. 
In the following, we first describe the experiment protocol and then provide results. 

\paragraph{Protocol} To implement the proposed method, we consider three types of users with different preferences for each of the financial goals:
\begin{itemize}
    \item the home buyer, whose priority is purchasing a home as quickly as possible;
    \item the saver/retirement planner, whose priority is maximizing retirement savings and saving for emergency;
    \item the debtor, who prefers to pay off debt first. 
\end{itemize}
For each category, we construct a representative user, with preference weights selected reflecting the user type (see Table~\ref{tab:profile}). 
\begin{table}[htbp]
\begin{center}
\begin{tabular}{l l}
\hline
\textbf{Users} & \textbf{Preference Weights}\\
\hline 
Home Buyer & \makecell[l]{Set $p^{\textrm{mortgage}} = 20.0$ \\
and other preference weights as one.}\\ \hline
Saver & \makecell[l]{Set
 $p^{\textrm{retirements}} =p^{401k} = q^{401k} = p^{IRA}= 20.0$,\\ 
 $p^{\textrm{emergency fund}}=5.0$, $q^{\textrm{emergency 
fund}} = 3.0$,\\
and other preference weights as one. }\\ \hline
Debtor &\makecell[l]{Set
$p^{\textrm{credit card}}=p^{\textrm{student loan}} = 20.0$, \\ $p^{\textrm{emergency fund}}=5.0$,
$q^{\textrm{emergency fund}}=3.0$,\\
and other preference weights as one.}\\ \hline
\end{tabular}
\caption {Preference weights for different users.} 
\label{tab:profile}  
\end{center}
\end{table}
For each of the users, we set the same input data (see Table~\ref{tab:inputs}). 
\begin{table}[htbp]
\begin{center}
\begin{tabular}{l l}
\hline
\textbf{Input parameter} & \textbf{Value}\\
\hline 
Monthly income & $\$7,500$\\ 
Inflation rate & $2\%$ \\
Stock Market Rate of Return & $10\%$\\
Credit Card Debt & $\$825$\\ 
Credit Card APR & $ 20\% $\\ 
Student Loan Debt & $\$80,000$\\ 
Student Loan APR & $4\%$ \\
Mortgage Down Payment & $\$157,000$\\ 
Emergency Amount (I) & $\$1,800$\\ 
Emergency Amount (II) & $\$9,000$\\ 
Retirement Savings & $\$1,000,000$\\ 
IRA (monthly contribution) & $\$500$\\ 
401K (min and max contribution) & $6\% - 13\%$ of salary\\
Time-horizon & $10$ years\\
\hline
\end{tabular}
\caption {Inputs for each user.} 
\label{tab:inputs}  
\end{center}
\end{table}
With this experiment, we demonstrate that the proposed paycheck optimization framework is able to effectively address the preferences of users, while finishing each financial goal in an efficient manner. 

\begin{figure}[htbp]
    \centering
    \includegraphics[scale = 0.45]{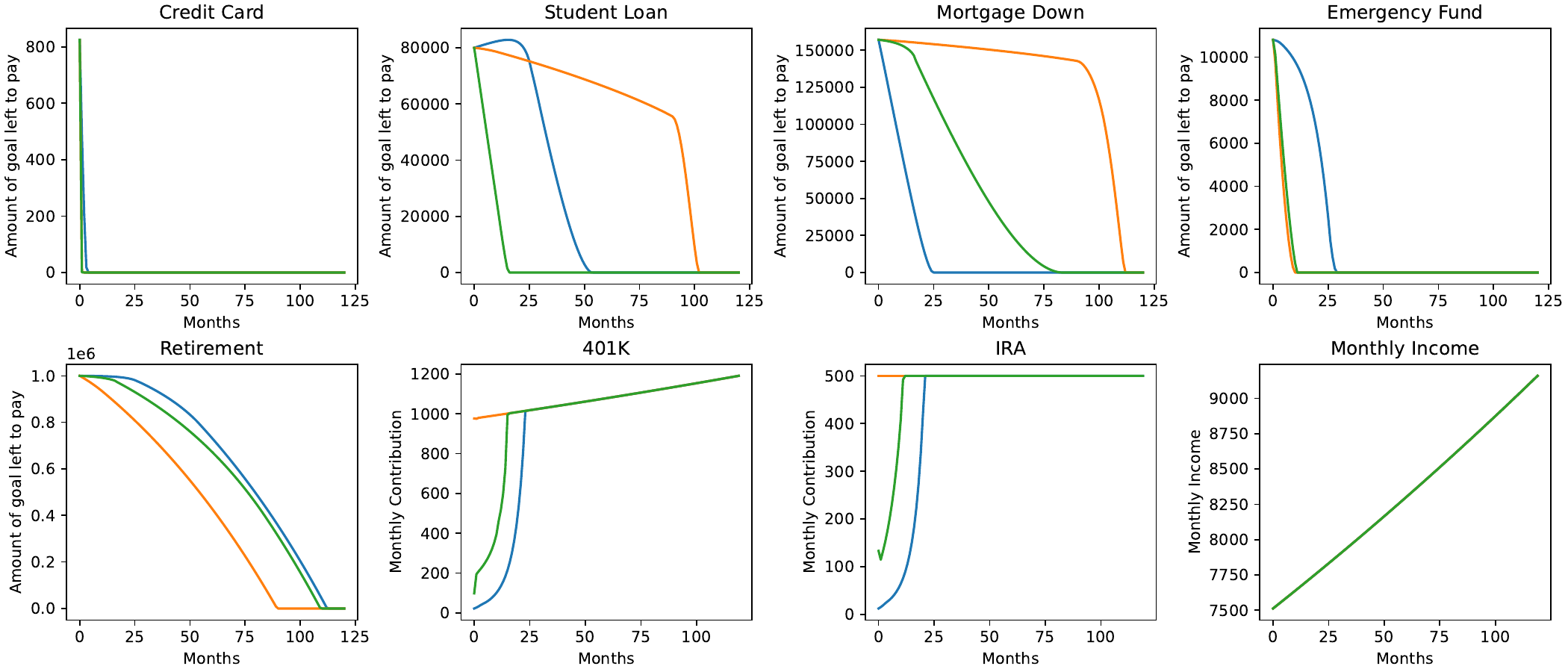}
    \caption{Contribution to each goal over time under the learned policy with constant rates for three different representative users: home buyer in blue, saver in orange, and debtor in green}
    \label{fig:constant_rates}
\end{figure}

\paragraph{Results} For each representative user, we report the contribution to each goal over time under the learned policy in Figure~\ref{fig:constant_rates}. 
We note that each goal is successfully completed under the learned paycheck allocation policy. 
Also, the result is consistent with the user preference. 
Specifically, home buyer does buy a mortgage earlier than others; the savings of saver grows faster; debtor pays off the debts in the fastest manner. 

\paragraph{Explainability}
From the experiments above, we can explain the learned policy by preference weights. 
Specifically, the policy function prefers to finish the financial goals with higher preference weights by allocating more income to such goals. 
One can further examine the effects of each state variable on the learned policy by using Shapley values.
Some examples in portfolio optimization include \citet{babaei2022explainable,colini2018variance}. 
However, it is nontrivial to extend such analysis from portfolio optimization to multiple financial goals in our setting.
We thus defer that to future work.

\section{Extension to Stochastic Rates}
\label{sec:extension}
In the previous analysis, we fixed the rates $\curly{r_t^i}_{i \in I}$ as constant over time, while in practice some or all the rates may evolve stochastically over time. 
In this section, we extend our method to the stochastic rate case. 
Specifically, we assume that the rates $\curly{r_t^i}_{i \in I}$ follow a Markov process, so that the dynamics of $X_t^i$ are also stochastic.
As a result, the value function in \eqref{eq:target} needs to be redefined as
\eq{target-2}{
V(\pi) = \EE_{\pi}[ \sum_{t=0}^{T}\sum_{i \in I} u_i(X_t^i) ],
}
where the expectation is over the dynamics of $X_t^i$ dependent on $r_t^i$, under the policy $\pi$.

We maximize the value function \eqref{eq:target-2} following the deep deterministic policy gradient in Section~\ref{sec:method} while using data of rates. 
We use $\tau=\curly{ r^i_{t}}_{t=0,i\in I}^T$ to denote a data trajectory of rates.  
Let $\curly{\tau_k}_{k=1}^n$ denote a dataset with $n$ observed trajectories.
We parameterize the policy function as deep neural network with parameter $\theta$:
\eq{policy-par}{
\pi(t) = f(x_t, \curly{r_t^i}_{i \in I}; \theta).
}
Then, we train the neural network $f$ by maximizing the sample-average utility function
\eq{average-utility}{
\frac{1}{n} \sum_{k=1}^n \sum_{t=0}^n\sum_{i\in I}u_i(x_t^{i,\theta, k}),
}
where $x_t^{i,\theta, k}$ denotes the state value at time point $t$ under the policy function \eqref{eq:policy-par} with stochastic rates following $\tau_k$. 
Thus, the gradient of \eqref{eq:average-utility}, with respect to $\theta$, is derived as
\eqs{
 \frac{1}{n}\sum_{k=1}^n \sum_{t=0}^{T}\sum_{i \in I} \frac{d u_i(x_t^{i,\theta,k})}{d\theta}.
}
In other words, in each iteration of our policy learning, we use the average over $n$ trajectories to calculate the gradient and update $\theta$.
Note that our procedure does not need independence assumptions or parametric models for the stochastic rates $\curly{r_t^i}_{i \in I}$: the dynamics of $\curly{r_t^i}_{i \in I}$ is purely handled by data.

\section{Simulation with Stochastic Rates}
\begin{figure}[htb]
    \centering
    \includegraphics[scale = 0.45]{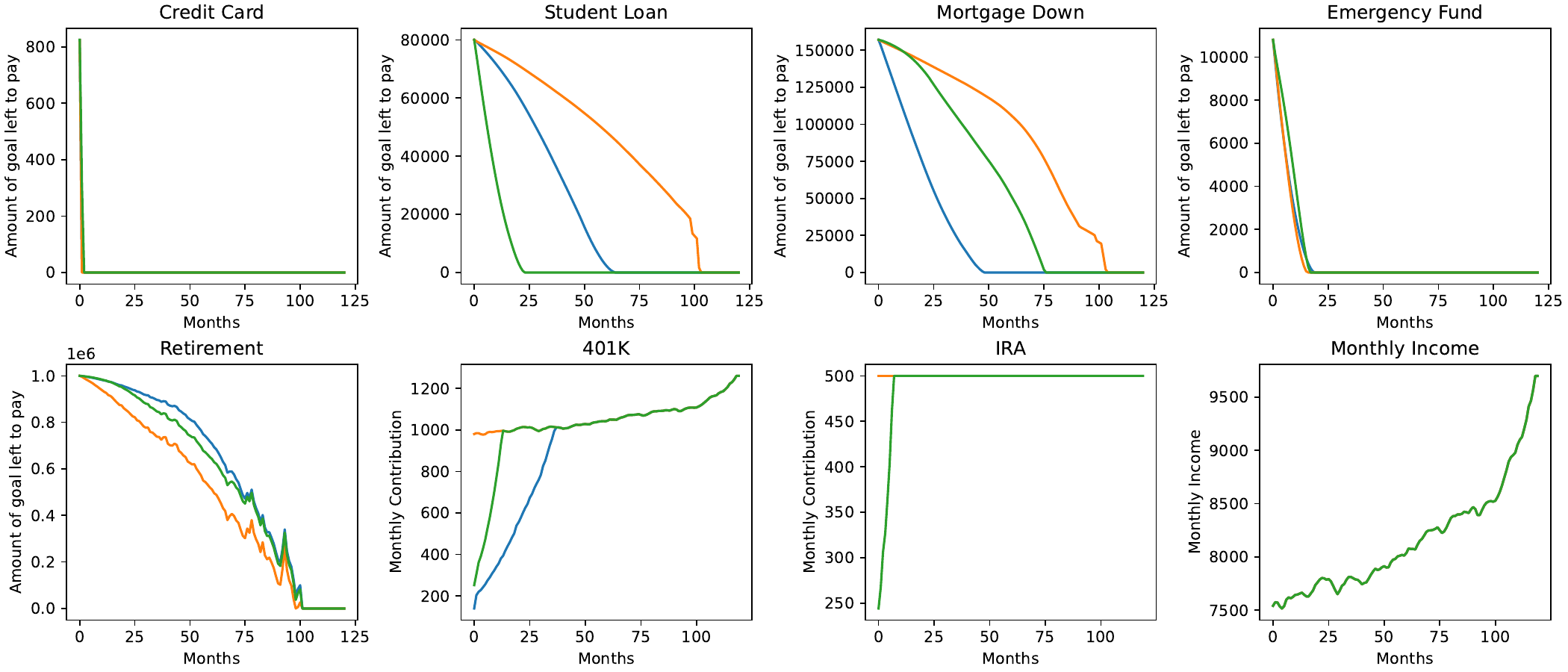}
    \caption{Contribution to each goal over time under the learned policy with stochastic rates for three different representative users: home buyer in blue, saver in orange, and debtor in green. \\ Note that the monthly income suffers a sharp increase after month $100$, since it is directly affected by inflation (which has  hiked over the last couple of years).}
    \label{fig:stochastic_rates}
\end{figure}

In this section, we conduct experiments for the case with stochastic rates. 
Under the same setup as the experiments in Section~\ref{sec:exp-deter}, we treat rates $r_t^i$ as stochastic processes instead of constants.
Following the procedure in Section~\ref{sec:extension}, to handle gradient estimation for \eqref{eq:target-2}, we select the Consumer Price Index (CPI), the market yield of 3-Month U.S. Treasury Bills and the S\&P500 Index return from 1985-2022 in order to model inflation, as well as the rate of return on mortgage down payment savings and retirement savings respectively.
Under the learned policy, the contributions over time for different users are reported in Figure~\ref{fig:stochastic_rates} over a ten year time horizon from 2012-2022.
Note that the results are mainly consistent with the deterministic rates case in Section~\ref{sec:exp-deter}.
However, here the contributions to each goal have more fluctuations, since the policy is implicitly trying to predict the potential rate change and adjust the paycheck assignment accordingly. 

\section{Conclusion}
We propose a framework for paycheck optimization with an end-to-end reinforcement learning solution.
By formalizing the problem into a piecewise linear utility maximization problem, our method is able to handle heterogeneous financial goals, the preferences of users, and also the stochastic rates. 
We empirically demonstrate the applicability of the proposed method. 

\appendix
\section{Example of waterfall failure}\label{app_waterfall}
Consider an individual with disposable income of $\$1000$ a month and two financial goals, each of which involve paying off debt:

\begin{itemize}
    \item Goal 1 is to pay off \$1000 with no interest rate, and with priority $p^1=1000$.
    \item Goal 2 is to pay off $\$\frac{1,000,000}{1+r}$ with interest rate $r=0.001$ and priority $p^2=1$.
\end{itemize}
Recall that the waterfall method consists in paying off the goals in order of priority. Accordingly, the strategy would be as follows.  First, we would pay off Goal 1, since $p^1>p^2$.  Hence, $X_0^1=1000$, $X_t^1=0~\forall t>0$.  Notice that, for goal $2$ we have $X_0^2=\frac{1,000,000}{1+r}$, while for goal $1$, $X_1^2=1,000,000$.  For each subsequent time $t>1$, notice that the increase in $X^2$ due to interest will be equal to $1000$.  Hence, at each time step $t$, the debt for Goal 2 will increase by 1000, resulting in the user needing to allocate all her  paycheck to pay down this increase.  Thus, $X_t^2=1,000,000$ for $t>0$, and the user will never be able to pay down Goal 2. 


On the other hand, let us consider a strategy where we recognize the threat of future compounding interest. The user would optimally split her paycheck evenly among the two goals.  Therefore, $X_0^1=1000,X_1^1=500,X_t^1=0$ for $t\geq 2$.  For Goal 2, $X_0^2=\frac{1,000,000}{1+r},X_1^2=999,500,X_2^2=999,999.5$.  From this point onward, the user would allocate all her paycheck, equal to $\$1000$ towards Goal 2, which will be gradually paid down. Indeed, the increase in the debt due to interest will always be smaller than her paycheck.

\section{Future Directions}
We envision the following directions as future work.
\begin{itemize}
    \item \emph{Portfolio Optimization.} The current paycheck optimization framework does not consider portfolio optimization. 
    However, portfolio optimization an important option for paycheck investment. 
    Therefore, as one future direction, we can incorporate portfolio optimization methods~\citep{giudici2022network,d2019promises,d2021robo, geng2023factor} into the current framework.     
    \item \emph{Healthcare.} In this work, we propose a novel decision-making framework that is able to unify different goals while incorporating user preferences regarding the goals. 
    This framework can be extended to tasks other than paycheck optimization. 
    One example is healthcare, where doctors often face decision-making problems like disease diagnosis~\citep{geng2020joint,geng2019partially}, drug selection~\citep{geng2018temporal,kuang2017screening}, and treatment selection~\citep{geng2019parathyroid}.  
    \item \emph{Offline RL and Inverse RL.} In the current framework, we leverage the deterministic policy gradient descent method.
    This method is an online RL method in the sense that it needs to interact with the environment in each iteration. 
    As a result, we construct a simulated dynamics in Section~\ref{sec:update}. 
    One methodological improvement can be to relax this modeling assumption by using offline RL or inverse RL methods~\citep{levine2020offline,geng2023improving,geng2020deep,geng2023data,arora2021survey}
\end{itemize}

\bibliography{paycheck}
\bibliographystyle{ref}

\end{document}

%% file: math_commands.tex
\usepackage{amsmath,amsfonts,bm}

\def\eqref#1{(\ref{#1})}

\def\1{\bm{1}}

